\documentclass[runningheads,a4paper]{llncs}

\usepackage{graphicx}
\usepackage{epstopdf}
\usepackage{amssymb}
\usepackage{booktabs}
\usepackage{verbatim}
\usepackage{algorithm}
\usepackage{algpseudocode}
\usepackage{xspace}
\usepackage{caption}
\usepackage[english]{babel}
\usepackage[utf8x]{inputenc}
\usepackage[T1]{fontenc}
\usepackage[table,xcdraw]{xcolor}
\usepackage{subfig}  
\usepackage{authblk}

\usepackage{rotating}
\usepackage{verbatimbox}

\usepackage{amsmath}
\usepackage{graphicx}
\usepackage[colorinlistoftodos]{todonotes}
\usepackage[colorlinks=true, allcolors=blue]{hyperref}
\usepackage{cite}

\date{}

\usepackage[overlay,absolute]{textpos}

\begin{document}

\begin{textblock*}{10in}(38mm, 10mm)
{\textbf{Ref:} \emph{International Conference on Artificial Neural Networks (ICANN)}, Springer LNCS,}
\end{textblock*}
\begin{textblock*}{10in}(38mm, 15mm)
{Vol.~11141, pp.~604--612, Rhodes, Greece, October 2018.}
\end{textblock*}

\title{DeepEthnic: Multi-Label Ethnic Classification \\
from Face Images}

\author{Katia Huri\inst{1} \and Eli (Omid) David\inst{1} \and Nathan S. Netanyahu\inst{1,2} }

\authorrunning{K.~Huri, E.O.~David, and N.S.~Netanyahu}

\institute{
Department of Computer Science, Bar-Ilan University, Ramat-Gan 5290002, Israel \\
\email{katiahuri@gmail.com, mail@elidavid.com, nathan@cs.biu.ac.il}
\and
Center for Automation Research, University of Maryland, College Park, MD 20742\\
\email{nathan@cfar.umd.edu}
}

\maketitle

\begin{abstract}
Ethnic group classification is a well-researched problem, which has been pursued mainly during the past two decades via traditional approaches of image processing and machine learning. In this paper, we propose a method of classifying an image face into an ethnic group by applying \textit{transfer learning} from a previously trained classification network for large-scale data recognition. 
Our proposed method yields state-of-the-art success rates of 99.02\%, 99.76\%, 99.2\%, and 96.7\%, respectively, for the four ethnic groups: African, Asian, Caucasian, and Indian.
\end{abstract}

\section{Introduction}
Ethnic classification from facial images has been studied for the past two decades with the purpose of understanding how humans perceive and determine an ethnic group from a given image. The motivation stems, for example, from the fact that (gender and) ethnicity play an important role in face-related applications, such as advertising, social insensitive-based systems, etc. Furthermore, while facial features are subject to change (due to aging, for example), ethnicity is of interest due to its invariance over time.

Recent works on demographic classification are divided conceptually into appearancebased methods (using, e.g., eigenface methods, fisherface methods, etc.) and geometry-based methods (relying, e.g., on geometric parameters, such as the distance between the eyes, face width and length, nose thickness, etc.). One of the main challenges of automatic demographic classification is to avoid any ``noise'', such as illumination, background distortion, and a subject’s pose.

In this paper, we introduce a deep learning-based method, that achieves state-of-the-art results for facial image representations and classification for the four ethnic groups: African, Asian, Caucasian, and Indian.

\section{Related Work}
\subsection{Traditional ML-Based Techniques}

During the past two decades, there has been enormous progress on the topic of ethnic group classification, using various classical Machine Learning methods. These approaches are based mainly on feature extraction and training classifiers; see Table \ref{traditionalMethods} below.

Hosoi et al.~\cite{Hosoi} were among the first to achieve promising results. They employed Gabor wavelet transformations for extracting key facial features, and then applied SVM classification. They reported classification accuracies of 94.3\%, 96.3\%, and 93.1\%, respectively, for the three ethnic groups: African, Asian, and Caucasian.

\vspace{-0.3cm}

\begin{table}[]
\centering
\caption{Previous work on ethnic group classification using traditional Machine Learning methods}
\resizebox{\textwidth}{!}{%
\begin{tabular}{|c||c|c|c||c|}
\hline
\rowcolor[HTML]{EFEFEF} 
\textbf{\begin{tabular}[c]{@{}c@{}}Authors \end{tabular}}              & \textbf{Approaches}                                                                                                  & \textbf{Databases}                                                                                           & \textbf{Ethnic groups}    & \textbf{Success rate}    \\ \hline \hline
\begin{tabular}[c]{@{}c@{}}Hosoi et al.\\ 2004~\cite{Hosoi}\end{tabular}                       & \begin{tabular}[c]{@{}c@{}}Gabor Wavelet \\ and SVM\end{tabular}                                                     & 1,991 face photos                                                                                             & African, Asian, Caucasian & 94.3\%, 96.3\%, 93.1\% \\ \hline
\begin{tabular}[c]{@{}c@{}}Lu et al.\\ 2004~\cite{XiaoguangLu}\end{tabular}                          & LDA                                                                                                                  & \begin{tabular}[c]{@{}c@{}}Union of DB\\ (2,630 photos of \\ 263 objects)\end{tabular}                         & Asian, non-Asian        & 96.3\% (Avg)             \\ \hline
\begin{tabular}[c]{@{}c@{}}Yang et al. \\ 2010~\cite{Yang}\end{tabular}                       & \begin{tabular}[c]{@{}c@{}}Real Adaboost\\ (Haar, LBPH)\end{tabular}                                                & \begin{tabular}[c]{@{}c@{}}FERET and PIE\\  (11,680 Asian and \\ 1,016
non-Asian)\end{tabular}                 & Asian, non-Asian        & 92.1\%, 93.2\%         \\ \hline
\begin{tabular}[c]{@{}c@{}}Lyle et al. \\ 2010~\cite{Lyle}\end{tabular}                       & \begin{tabular}[c]{@{}c@{}}Perioucular regions,\\ LBP, SVM\end{tabular}                                              & \begin{tabular}[c]{@{}c@{}}FRGC \\ (4,232 faces, 404 objects)\end{tabular}                                    & Asian, non-Asian        & 92\% (Avg)             \\ \hline
\begin{tabular}[c]{@{}c@{}}Guo et al.\\ 2010~\cite{Guo}\end{tabular}                         & \begin{tabular}[c]{@{}c@{}}Biologically inspired\\ features\end{tabular}                                             & \begin{tabular}[c]{@{}c@{}}MORPH-II \\ (10,530 Africans, 10,530 Caucasians)\end{tabular}                               & African, Caucasian           & 99.1\% (Avg)             \\ \hline
\begin{tabular}[c]{@{}c@{}}Xie et al.\\ 2012~\cite{Xie}\end{tabular}                         & \begin{tabular}[c]{@{}c@{}}Kernal class dependent \\ feature analysis\\ (KCFA)\end{tabular}                          & \begin{tabular}[c]{@{}c@{}}MBGC DB \\ (10,000 African,\\ 10,000 Asian, 20,000 Caucasian)\end{tabular}           & African, Asian, Caucasian  & 97\%, 95\%, 97\%       \\ \hline
\end{tabular}
}
\label{traditionalMethods}
\end{table}
\vspace{-0.3cm}

Lu et. al~\cite{XiaoguangLu} constructed an \textit{ensemble framework}, which integrates LDA applied to the input face images at different scales. The combination strategy in the ensemble is the product rule~\cite{Kittler} to combine the outputs of individual classifiers at these different scales. Their binary classifier of Asian and non-Asian classes obtained success rates of 96.3\%, on average.

Yang et al.~\cite{Yang} used LBPH\footnote{LBPH is a combination of \emph{local binary pattern} (LBP) with the \emph{histogram of oriented gradients} (HOG) techniques.}~\cite{HOG} to extract features of texture descriptions, in order to enhance considerably the human detection algorithm that was previously suggested by Xiaoyu et al.~\cite{Xiaoyu}. Real AdaBoost was then used iteratively to learn a sequence of best local features to create a strong classifier. Their binary classifier of Asian and non-Asian classes had success rates of 92.1\% and 93.2\%, respectively.

Lyle et al.~\cite{Lyle} extracted ethnicity information from the \textit{periocular region images}\footnote{A periocular region includes the iris, eyes, eyelids, eye lashes, and part of the eyebrows.} using grayscale pixel intensities and periocular texture features computed by LBP. Their binary SVM classifier of Asian and non-Asian classes yields success rates of 93\% and 91\%, respectively,

Guo et al.~\cite{Guo} proposed using \textit{biologically-inspired features} for ethnic classification, by applying a battery of linear filters to an image and using the filtered images as primary features~\cite{Jarrett}). Their binary classifier to Africans and Caucasians achieved 99.1\% success rate, on average. However, integrating the three ethnic groups: Asian, Hispanic, and Indian, result in a sharp success rate decrease. Specifically, the accuracies recorded were
African: 98.3\%, Caucasian: 97.1\%, Hispanic: 59.5\%, Asian: 74.2\%, and Indian: 6.9\%.

Xie et al.~\cite{Xie} used \textit{kernel class-dependent feature analysis} for generating nonlinear features (by mapping them onto a higher-dimensional feature space which allows higher order correlations~\cite{XieChunyan}) and facial color-based features to classify large-scale face databases. Their classifier achieved success rates of 97\%, 95\%, and 97\%, respectively, for the three ethnic groups: African, Asian, and Caucasian.

To summarize, although some of the surveyed methods yield  high classification results, it appears that they are limited to laboratory conditions, i.e., they may not perform as well on a diverse, large-scale database, consisting of face images of different gender, pose, age, illumination conditions, etc. In contrast, we create in this work a diverse face image database for training and testing.

\subsection{Recent Deep Learning Techniques}

Ethnic group classification has improved significantly in recent years, due to the use of deep learning ( techniques, e.g., CNN architectures, enhanced feature extraction, etc. (See Table ~\ref{previousDL} for an overview.)

\vspace{-0.3cm}
\begin{table}[]
\centering
\caption{DL-based methods for ethnic group classification}
\resizebox{\textwidth}{!}{%
\begin{tabular}{|c||c|c|c||c|}
\hline 
\rowcolor[HTML]{EFEFEF} 
\textbf{Authors}                                                                  & \textbf{Approach}                                                                                                           & \textbf{Databases}                   & \textbf{Ethnic groups}                                                                                              & \textbf{Success rate}                                                                    \\ \hline \hline
\begin{tabular}[c]{@{}c@{}}Ahmed et al.~\cite{Ahmed} \\ 2008 \end{tabular} & \multicolumn{1}{l|}{\begin{tabular}[c]{@{}c@{}}Transfer learning from\\pseudo tasks\\ (CNN + transfer learning)\end{tabular}} & \multicolumn{1}{l|}{FRGC 2.0, FERET} & \multicolumn{1}{l||}{Asian, Caucasian, Other}                                                                            & \multicolumn{1}{c|}{95.4\% (avg.)}                                                      \\ \hline
\begin{tabular}[c]{@{}c@{}}Inzamam et al.~\cite{Inzamam}\\ 2017 \end{tabular}                      & \begin{tabular}[c]{@{}c@{}}Feature extraction due to ANN\\ and SVM classification\end{tabular}                                     & 10 different DBs                      & African, Asian, Caucasian                                                                                           &  99.66\%, 98.28\%, 99.05\%                                                              \\ \hline
\begin{tabular}[c]{@{}c@{}}Wang et al.~\cite{WeiWang}\\ 2017 \end{tabular}                       & CNN                                                                                                                           & Varaiety of DBs                    & \begin{tabular}[c]{@{}c@{}}African, Caucasian\\ Chinese, non-Chinese\\ Han, Uyghur, non-Chinese \end{tabular} & \begin{tabular}[c]{@{}c@{}} 99.4\%, 100\%\\  99.62\%, 99.38\%\\ 99\% (avg.) \end{tabular} \\ \hline 
\end{tabular}
}
\label{previousDL}
\end{table}
\vspace{-0.3cm}

Ahmed et al.~\cite{Ahmed} were the first to apply transfer learning for ethnic classification. Their classifier achieved a success rate of 95.4\%, on average, for the ethnic groups: Asian, Caucasian and ``Other'', using the FRGC 2.0 and FERET databases for training data.

Inzamam et al.~\cite{Inzamam} performed the classification by extracting features from a deep neural network followed by SVM classification on 10 datasets (13,394 images in total, including different variations of the FERET, CASPEAL, and Yale databases). Their classifier achieved success rates of 99.66\%, 98.28\%, and 99.05\%, respectively, for the ethnic groups: African, Asian, and Caucasian.

Wang et al.~\cite{WeiWang} used deep CNNs to extract features and classify them simultaneously. Three different classifiers were created: (1) The first binary classifier for African and Caucasian classes, achieving success rates of 99.4\% and 100\%, respectively; (2) a binary classifier for Chinese and non-Chinese classes, achieving success rates of 99.62\% and 99.38\%, respectively; and (3) a 3-way classifier for Han, Uyghur, and non-Chinese classes, achieving an average success rate of 99\%.

\section{Proposed Method}
\subsection{Data Source}
As previously indicated, the purpose of this research is to distinguish between the four ethnic groups: African. Asian, Caucasian, and Indian. We created our dataset by combining 10 different databases, originally proposed for the problem of \emph{face recognition}, and then sorting them into the ethnic groups of interest. The databases included IMFDB~\cite{IMFDB}, CNBC~\cite{CNBC}, Labeled Faces in the Wild (LFW)~\cite{LFW}, the Essex face dataset~\cite{ESSEX}, Face Tracer~\cite{FaceTracer}, the Yale face database ~\cite{YALE}, SCUT5000~\cite{SCUT5000}, and additional collected image datasets. We also used the well-known FERET database~\cite{FERET1,FERET2}, which contains facial images collected under the FERET program, sponsored at the time by the U.S. Department of Defense (DoD). 

Altogether, the collected dataset contains images of various sizes.

\subsection{Facial Image Preprocessing}
As part of preprocessing, the data should be normalized to be compatible with the network's architecture. Also, it is denoised to make it as clean as possible. Thus, we first convert every RGB image to a grayscale one to create a homogeneous dataset of grayscale images.

Note that our collection also contains datasets (such as AT\&T and CASPEAL) of only grayscale face images.
We then use the Face Cascade detector (part of the OpenCV library), to detect and crop the faces. After detecting and cropping the faces, the cropped images are downscaled to $80 \times 80$ pixels and are denoised using a non-local means denoising algorithm (implemented by the OpenCV function, fastNlMeansDenoising).

Finally, (grayscale) images are duplicated to create an image size of
$80 \times 80 \times 3$. This is done to be compatible with the VGG-16 network, which receives three-channel images as input. 

\begin{figure}[]
\centering
\small
\includegraphics[width=0.3\textwidth]{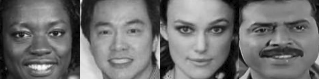}
\caption{\label{fig:EthnicGroups} Examples of preprocessed face images and their ethnic group labels (from left to right): African, Asian, Caucasian, and Indian.}
\end{figure}

After preprocessing, the face images were sorted manually into the four ethnic groups of interest (i.e., African, Asian, Caucasian, and Indian), creating a labeled face database for ethnic group classification. See Fig.~\ref{fig:EthnicGroups} for specific face images per each ethnic group. 

\vspace{-0.3cm}

Since the number of images acquired was rather imbalanced over the four ethnic groups, we perturbed each image in the smaller training samples with a minor Gaussian noise, so as to augment these training samples with slightly different duplicates.

\subsection{Transfer Learning}\label{TransLearning}
Due to the challenging problems DL has to solve, it takes enormous resources (mostly training time, but also fast computers, training data storage, and human expertise) to train such models.

Transfer learning is an ML technique that helps to overcome those issues by using a model that was trained on a specific task (without any changes to the weights) to solve other tasks.

Yosinski et al.~\cite{Yosinski} showed that using transfer learning can solve all of these issues and create more efficient and accurate models for solving additional problems. It is important to note that transfer learning only works if the model features learned from the first task are sufficiently generic.

A pretrained model has been previously trained on a dataset and contains the weights and biases that represent the features of the data it has seen during training. The most commonly used pre-trained models are VGG16, VGG19~\cite{Simonyan}, and Inception V3~\cite{Szegedy}, due to their high success rate and improvement on the ImageNet dataset classification problem. 

VGG16 is a classification model with 16 layers, which is based on the ImageNet dataset and can classify 1,000 different image types (including animals, buildings, and humans). The model's weight file size is 528 MB, and it can be easily accessed for free.

\subsection{Network Architecture}

\begin{figure}[]
\centering
\subfloat[]{\label{origVG16}\includegraphics[height= 3cm, width=0.8\textwidth]{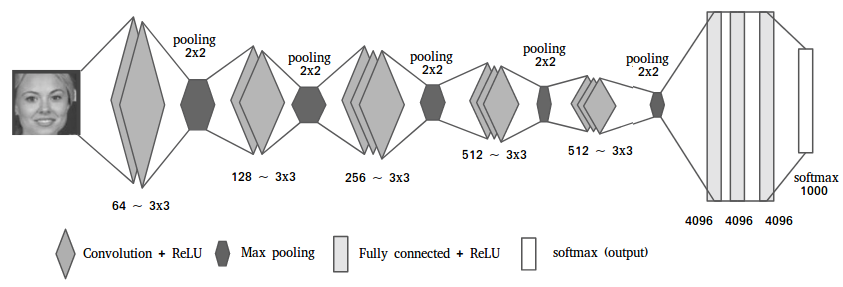}} \\
\subfloat[]{\label{TransferVGG16}\includegraphics[height=3cm, width=0.8\textwidth]{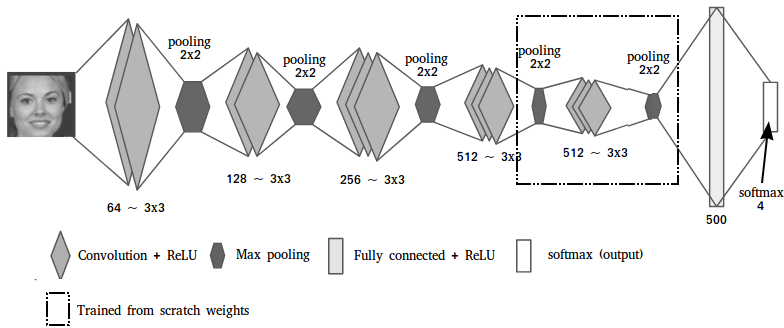}}
\caption{(a) Original version of VGG-16, and (b) our modified architecture for partial transfer learning from VGG-16.}
\label{fig:TransferNetwork}
\end{figure}

Fig.~\ref{fig:TransferNetwork} shows the original VGG-16 architecture and our modified architecture, which inputs a preprocessed $80 \times 80 \times 3$ image and outputs its predicted ethnic group.

The modified architecture contains the original, previously trained VGG-16 network, without the final three fully-connected layers and the original softmax layer.

The five layers of the remained network (i.e., max pooling layer, three convolution layers and activation layers, and the final max pooling layer) were selected after experimenting extensively with a large number of possibilities for retraining the entire network. Note that running on the original network ``as-is'' would have given very poor results. Instead, the idea is to capture universal features (like curves and edges), and further refine these features due to the above modified layers, by retraining the entire network in the context of our problem. 

Specifically, the classification softmax layer was replaced by a fully-connected layer of size $n = 500$ and a softmax layer (of size $p = 4$), which outputs a probability distribution for the ethnic group classification.

The output of the softmax layer is the probability distribution for the classification problem. We train the network with the purpose of minimizing the cross-entropy loss function. The network is trained using \textit {stochastic gradient descent} (SGD), as part of the backpropagation phase.

\section{Experiments and Results}
We present the datasets used in the experiments, and give detailed empirical results of the 10-fold cross validation for the four-class ethnic group classification.

To increase the classification success rate, we first experimented with different network hyper-parameters, e.g., number of epochs, type of activation function, size of the fully-connected layer to add, type of loss function, etc. After running a grid search on the hyper parameters options, we selected the following set of hyper-parameters, which provided the best performance: 50 epochs, a ReLu~\cite{Hinton2} activation function (an element-wise operation applied per pixel), an additional fully-connected layer of 500 neurons, and a categorical cross-entropy loss function.

We trained and tested the model for each fold, by allocating each time 75\%, 10\%, and 15\% of the data, respectively, to training, validation, and testing.

The training time using TensorFlow and Keras infrastructure on GeForce GTX 1070 was roughly 4.5 hours (compared to nearly 11.5 hours for training from scratch on the same architecture), and the real-time evaluation of an image is about 10 msec.

We ran a 10-fold cross validation using the selected base model, and obtained  classification accuracies of 99.02\%, 99.76\%, 99.18\%, and 96.72\%, respectively, for the categories, African, Asian, Caucasian, and Indian. Bottom-line accuracies and loss for the Ethnic classes are summarized in Table~\ref{transferLearning10FoldFinal}.

\vspace{-0.5cm}
\begin{table}[]
\centering
\caption{Summary of total success rate and loss over entire experiments}
\begin{tabular}{|c|c|c|c||c|l|}
\hline 
\rowcolor[HTML]{EFEFEF} 
African & Asian   & Caucasian & Indian  & Total Success rate & Total Loss \\ \hline \hline
99.02\% & 99.76\% & 99.18\%   & 96.72\% & 99.18\%        & 0.03518    \\ \hline 
\end{tabular}
\label{transferLearning10FoldFinal}
\end{table}

\section{Conclusions}
In this paper we presented a novel approach to the ethnic group classification problem. By modifying a previously trained classification network (namely VGG-16) for transfer learning we achieved state-of-the-art performance with respect to four ethnic classes: African, Asian, Caucasian, and Indian. Specifically, we obtained higher success rate levels for a larger number of classes, while working with a more diverse dataset than previously reported. Also, our derived scheme exhibits faster training time then training from scratch with similar results. 

Our future work will focus on extending the number of classes, and improving the robustness of the proposed method to different image conditions, such as different head poses, illumination change, etc. 

\bibliography{./deepethnic} 
\bibliographystyle{splncs03}

\end{document}